\documentclass[pmlr]{jmlr}

\RequirePackage{graphicx}
 \usepackage{booktabs}
\usepackage{makecell}
\usepackage{wrapfig}
\usepackage{floatrow} 
\usepackage{longtable}
\usepackage{amssymb}
\usepackage{bbm}
\usepackage{authblk}

\makeatletter
\def\set@curr@file#1{\def\@curr@file{#1}} 
\makeatother
\usepackage[load-configurations=version-1]{siunitx} 

\theorembodyfont{\upshape}
\theoremheaderfont{\scshape}
\theorempostheader{:}
\theoremsep{\newline}

\jmlrvolume{}
\jmlryear{2024}
\jmlrworkshop{ }

\title[Short Title]{Direct Preference Optimization for Suppressing Hallucinated Prior Exams in Radiology Report Generation}

\author[1]{Oishi Banerjee}
\author[2]{Hong-Yu Zhou}
\author[3]{Subathra Adithan}
\author[4]{Stephen Kwak}
\author[2]{Kay Wu}
\author[2]{Pranav Rajpurkar}

\affil[1]{Department of Computer Science, Harvard University}
\affil[2]{Department of Biomedical Informatics, Harvard University}
\affil[3]{Department of Radiodiagnosis, Jawaharlal Institute of Postgraduate Medical Education \& Research}
\affil[4]{Department of Radiology, Johns Hopkins University}

\begin{document}
\maketitle

\begin{abstract}

Recent advances in generative vision-language models (VLMs) have exciting potential implications for AI in radiology, yet VLMs are also known to produce hallucinations, nonsensical text, and other unwanted behaviors that can waste clinicians' time and cause patient harm. Drawing on recent work on direct preference optimization (DPO), we propose a simple method for modifying the behavior of pretrained VLMs performing radiology report generation by suppressing unwanted types of generations. We apply our method to the prevention of hallucinations of prior exams, addressing a long-established problem behavior in models performing chest X-ray report generation. Across our experiments, we find that DPO fine-tuning achieves a 3.2-4.8x reduction in lines hallucinating prior exams while maintaining model performance on clinical accuracy metrics. Our work is, to the best of our knowledge, the first work to apply DPO to medical VLMs, providing a data- and compute- efficient way to suppress problem behaviors while maintaining overall clinical accuracy.

\end{abstract}

\section{Introduction}

Recent advances in generative vision-language models (VLMs) have exciting potential implications for AI in radiology. Typically pretrained through self-supervised learning on large image-text datasets, VLMs can achieve high performance on complex multimodal tasks, interpreting radiology images and describing them in sophisticated medical text. Unfortunately VLMs are also known to produce hallucinations, nonsensical text, and other unwanted behaviors associated with large generative AI models. In the context of radiology, such behaviors are likely to increase clinician workloads, forcing them to spend time searching for and fixing AI errors, and may lead to patient harm if mistakes are left uncorrected.

To address these challenges, multiple techniques for reducing errors in generative models have been proposed. One approach is to preemptively prevent unwanted behaviors by carefully modifying pretraining datasets, such as by removing references to an off-limits topic. However, preprocessing large pretraining datasets in this fashion can become prohibitively expensive or time-consuming; furthermore, modifications to pretraining data cannot resolve issues found later on, after pretraining is already complete. In contrast, methods based on reinforcement learning with human or AI feedback (RLHF) can be applied after pretraining, improving already strong models by making their outputs better align with human preferences. Direct preference optimization (DPO), a technique that recently grew out of RLHF, offers a particularly simple and stable way to reduce unwanted behaviors in pretrained models, as it dispenses with the explicit reward model typically needed for reinforcement learning \citep{rafailov2023direct}. RLHF-based methods have shown considerable promise in the general domain, but they are not yet well-explored in the context of medicine.

In this paper, we propose and validate a DPO-inspired method for suppressing unwanted types of generations from VLMs performing chest X-ray (CXR) report generation. Radiology report generation is a challenging task for VLMs, requiring them to identify all diseases or other features of interest in a radiology image and fluently describe them in a short essay, ideally while taking into account the particular patient's history. We specifically apply our method to reducing the hallucination of prior exams, which has long raised concerns in radiology report generation \citep{hyland2024maira1, evaluating-progress-patterns, ramesh2022improving, miura-etal-2021-improving}. When engaging in this unwanted behavior, models generate hallucinatory statements such as ``The lungs are hyperinflated with emphysematous changes as seen on prior CT'' or ``There has been interval increase in right-sided opacity [i.e. compared to a previous image],'' despite not having access to any prior exam. These hallucinations can reduce the utility of AI-generated radiology reports for multiple reasons. First, they require radiologists to spend extra effort checking and editing these references, which could prove misleading if left uncorrected. Furthermore, references to prior exams can reduce the amount of clinically useful information in a report; for example, a report that claims ``cardiac size is unchanged'' fails to make clear whether the heart size is abnormal or not. We show that our method makes a pretrained VLM substantially less likely to hallucinate prior exams, while simultaneously maintaining the model's clinical accuracy.

\section{Main Insights}
We make the following specific contributions:

\begin{enumerate}
    \item We introduce and evaluate several DPO methods for radiology report generation, including both standard and weighted DPO losses. We generally find that DPO methods can be used to selectively suppress unwanted behaviors with little impact on clinical accuracy.
    \item We create a subset of MIMIC-CXR's training set, where GPT-4 has removed the majority of references to prior exams from the findings and impression sections of each report. These can be used in future research on reducing references to prior exams.
    \item We also create subsets of MIMIC-CXR's validation and test sets, where GPT-4 has removed the majority of references to prior exams from the findings and impression sections. These may be broadly useful when evaluating models, potentially providing an alternative to MIMIC-CXR that better penalizes models for hallucinating prior exams.
    
\end{enumerate}

Our work is, to the best of our knowledge, the first work to apply direct preference optimization to medical VLMs, successfully suppressing a problematic behavior while maintaining overall clinical accuracy. Compared to typical pretraining processes, our method is relatively compute- and data-efficient, making it a practical option even under resource constraints.

\section{Related Work}

\textit{Generative AI for Radiology Report Generation.} In the past, some methods for AI-based radiology reporting have avoided training VLMs to produce free-text reports, relying on other approaches such as retrieval \citep{jeong2023multimodal, pmlr-v158-endo21a}, template-based reporting \citep{kale2023replace}, or the use of GPT to reformat outputs from image models in text \citep{yan-etal-2023-style, ranjit2023retrieval}. However, powerful VLMs that directly generate free-text radiology reports based on input chest X-rays are rapidly growing in popularity \citep{chaves2024training, tu2023generalist, xu2023elixr, hyland2024maira1}. Some works have used reinforcement learning to improve the performance of such generative models \citep{zhou2024large, qin-song-2022-reinforced, miura-etal-2021-improving}. These methods have historically required the construction of explicit reward models and focused on overall clinical accuracy, while our method does not require an extra reward model and aims to target and suppress more specific unwanted behaviors.\\

\noindent \textit{Direct Preference Optimization.} Direct preference optimization (DPO) simplifies RLHF by training generative models directly on preferred and dispreferred responses for a given prompt, without requiring a new reward model \citep{rafailov2023direct}. DPO can effectively suppress hallucinations \citep{castricato2024suppressing}, including in image captions generated by VLMs \citep{zhou2024aligning, gunjal2024detecting}. Importantly, DPO has been shown to align pretrained models with human preferences without compromising their existing domain knowledge, making them a promising choice for radiology report generation \citep{ivison2023camels}. While the standard DPO loss typically places equal importance on all parts of a preferred or dispreferred response, a few variants prioritize certain response pairs \citep{liu2024lipo,amini2024direct} or certain parts of an individual response \citep{gunjal2024detecting}; inspired by this work, we experiment with weighted DPO losses that prioritize certain parts of a response. A handful of works have used DPO to fine-tune medical LLMs \citep{ahn2024note,ye2023qilinmed,dou2024integrating}, but to the best of our knowledge, ours is the first DPO method for medical VLMs.

\section{Methods}

\subsection{Direct Preference Optimization}

To perform DPO, we require a reference model $\pi_{\text{ref}}$, which has already been pretrained for the overall task of radiology report generation. We also require a preference dataset, where every datapoint consists of a prompt $x$, a preferred response $y_w$, and a dispreferred response $y_l$. In the context of radiology report generation, a prompt might contain the image(s) from a study, as well as available background information such as the indication. The preferred response would be the desired report, while the dispreferred response would be an undesirable version of the report that demonstrates the unwanted behavior that should be suppressed. Given prompt $x$, the standard DPO loss essentially rewards the policy model, $\pi_\theta$, for assigning high probabilities to $y_w$ and low probabilities to $y_l$, relative to the reference model:
\begin{equation}
\mathcal{L_{\text{DPO}}} = -
\log \sigma\biggl(\beta  
\bigl(\log\pi_\theta(y_w|x) - \log\pi_{\text{ref}}(y_w|x)\bigr)\\
- \beta \bigl(\log\pi_\theta(y_l|x) - \log\pi_{\text{ref}}(y_l|x)\bigr)
\biggr)
\end{equation}
where $\sigma(\cdot)$ denotes the sigmoid function.

Notably, this standard DPO loss assigns equal weight to all tokens in each response, regardless of whether those tokens relate to the behavior we wish to suppress. This property may be useful when suppressing broad behaviors such as harmfulness, which may affect all parts of a dispreferred response. However, many clinical use cases for DPO will focus on suppressing far more specific behaviors, which affect only a small self-contained part of a dispreferred response. For example, in our use case, a dispreferred response may refer to a prior exam in one line and otherwise be identical to the preferred response. Concerningly, the standard DPO loss would harshly penalize our model for producing any of the tokens in $y_l$, even though most lines in $y_l$ contain unobjectionable, correct content.

To suppress specific behaviors without penalizing unrelated correct statements, we also explore weighted DPO losses that selectively pay extra attention to certain parts of each response. First, we assume every token in the responses is labeled as either relevant or irrelevant to our behavior of interest. We next introduce a new hyperparameter $\gamma$, where 0 \underline{$<$} $\gamma$ \underline{$<$} 1. When calculating $\log\pi(y|x)$, we multiply the log-probabilities of irrelevant tokens by $\gamma$, effectively reducing their importance in comparison to the relevant tokens. Setting $\gamma$ = 0 essentially skips irrelevant tokens, while $\gamma$ = 1 gives the standard DPO loss where all tokens are weighted equally. To revisit the previous example where $y_l$ contained one reference to a prior exam, a weighted DPO loss would allow us to penalize the one  problematic line in $y_l$ more heavily than the other prior-free lines. 

To formally define the loss, let us assume $y$ consists of substrings $y_1$ through $y_n$ and that there are corresponding binary variables $r_1$ through $r_n$, where $r_i$ = 1 if $y_i$ is relevant to the behavior of interest and 0 otherwise. The loss can be written as follows:

\begin{equation}
\mathcal{L_{\text{WDPO}}} = 
-\log \sigma\biggl(\beta  
\bigl( f(\pi_\theta,y_w,x) - f(\pi_{\text{ref}},y_w,x)\bigr)\\
- \beta \bigl(f(\pi_\theta,y_l,x) - f(\pi_{\text{ref}},y_l,x) \bigr)\\
\biggr),
\end{equation}
where the function $f$ provides the weighted log probability that a model $\pi$ will produce a response $y$, given a prompt $x$:
\begin{equation}
f(\pi, y, x) = \sum_{i=1}^{n} \bigl(\mathbbm{1}[r_i=1]+\gamma\mathbbm{1}[r_i=0]\bigr)\log\pi\bigl(y_i | (x, y_1 ... y_{i-1})\bigr),
\end{equation}
where $\mathbbm{1}(\cdot)$ stands for the indicator function.

\subsection{Pretrained Model}
We pretrain a VLM through next-token prediction on MIMIC-CXR~\citep{johnson2019mimic}. We use this pretrained model as the reference model when calculating DPO losses and as the initial model checkpoint for all our fine-tuning experiments. Additionally, we use this pretrained model as a baseline model and report its performance.

The VLM mainly consists of three components: a vision encoder, a vision-language adapter, and a LLM. The vision encoder first converts the input image into visual tokens, which are mapped to the language space by the adapter. The processed visual tokens are then forwarded through the LLM along with the background information and the prompt to generate the CXR report. Specifically, we use the base model of Swin Transformer~\citep{liu2021swin} as the vision encoder. The adapter design consists of three layers. The first layer reduces visual tokens to manage GPU memory using adaptive pooling. Then, layer normalization is applied, followed by a linear projection to map visual representations to the language space. We initialize the LLM using Llama2-Chat-7b~\citep{touvron2023llama} and apply parameter-efficient tuning using LoRA~\citep{hu2021lora}.

\subsection{Dataset Creation}
We build our datasets using CXRs and reports from MIMIC-CXR, drawing from MIMIC-CXR's standard train/validation/test split. In the training set, we include reports that contain a ``findings'' section and/or an ``impression'' section marked under clear headers (i.e. ``FINDINGS:'' and ``IMPRESSION:'', ``CONCLUSION:'' or ``SUMMARY:''). In the validation and test sets, we only include reports that contain both a ``findings'' section and an ``impression'' section marked under clear headers. Where available, we also extract the indication and comparison sections and include these with the image in our prompts.

\begin{figure}
\includegraphics[width=\textwidth]{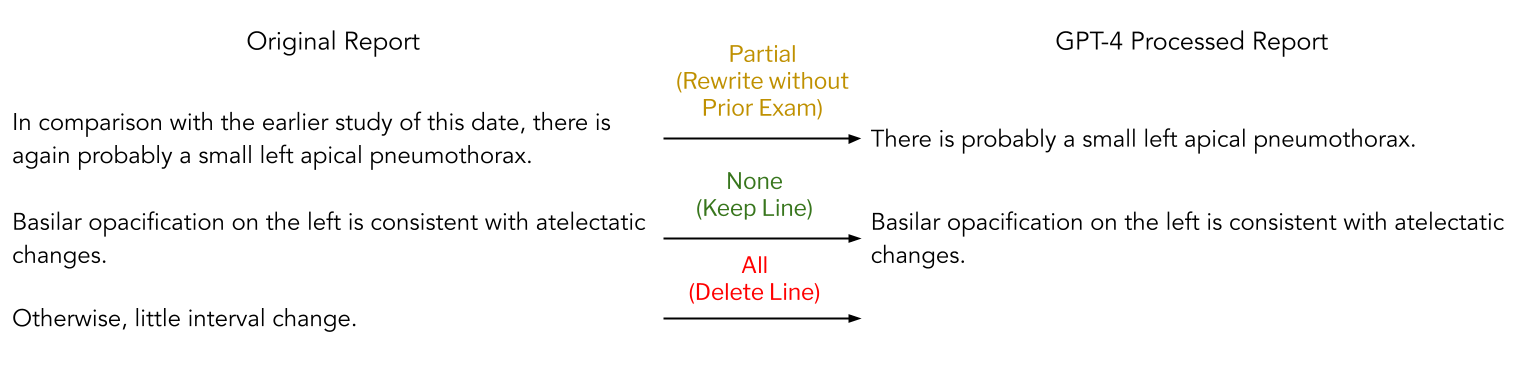}
\caption{GPT-4 labels each line in a report based on how much of it depends on a prior exam (``none'', ``partial'', ``all'') and rewrites ``partial'' lines to omit references to prior exams. }
\end{figure}

\subsection{Removing References to Prior Exams with GPT-4}
 We use GPT-4 to edit MIMIC-CXR's reports so they avoid references to prior exams, using few-shot prompting. In collaboration with radiologists, we developed a list of keywords often associated with references to prior exams, and we customize the examples in our few-shot prompt for each report, depending on which keywords are present. The prompt instructs GPT-4 to examine every line in the original MIMIC-CXR report and label how much of the line depends on references to prior exams: ``none'', ``partial'', or ``all''. For further details on the prompt, please see Appendix A. 
 Depending on the label, we take the following actions (Figure 1):
 \begin{enumerate}
    \item  If the label is ``none'', we copy the line into the edited report without making any changes. When calculating the weighted DPO loss, we treat this line as an ``irrelevant'' part of both the dispreferred and preferred responses. 
    \item If the label is ``partial'', we prompt GPT-4 to rewrite the line without references to prior exams, and we use the rewritten line in the edited report (e.g. ``Consolidation has worsened'' becomes ``Consolidation is present''). When calculating the weighted DPO loss, we treat the original line as a ``relevant'' part of the dispreferred response. In addition, we treat the rewritten line as a ``relevant'' part of the preferred response. 
    \item If the label is ``all'', we omit the line from the edited report. When calculating the weighted DPO loss, we treat this line as a ``relevant'' part of the dispreferred response; this line is absent from the preferred response. 
  \end{enumerate}

To preserve all possible information from the ground-truth report, we instruct GPT-4 to generally prefer the label of ``partial'' over ``all''. We reserve the label of ``all'' for extreme cases where a line offers no meaningful information without access to the prior exam (e.g. ``No change from prior image'', ``Cardiac size is unchanged''). Sometimes, a report's findings or impression section is entirely deleted when we remove references to prior exams because all its lines are labeled ``all''. If even one section is missing, either because it is removed during processing or because it is not found in the original report, we exclude the report from the validation and test datasets. If both sections end up missing, we exclude the report from the training datasets.

\subsection{Training Dataset}
We create a DPO training dataset, consisting of dispreferred reports that reference prior exams and preferred reports that avoid such references. To identify dispreferred reports, we identify 20,000 reports in MIMIC-CXR's training set. To improve dataset diversity, we select reports with unique values when we concatenate the findings and impression section. Additionally, we only select reports that contain the substring ``compar'' in the findings or impression section, as these reports are likely to contain at least one reference to a prior exam. To produce the preferred version of each report, we remove references to prior exams using GPT-4, as described above. We filter out 162 reports that receive malformed responses from GPT-4 (e.g. outputs for hallucinated extra lines, invalid JSON syntax). After removing reports where both the findings and impression section are missing, we arrive at a final training dataset with 19,806 studies.

\subsection{Validation Dataset}
To produce our validation dataset, we use GPT-4 to remove references to prior exams from 1,718 reports in the MIMIC-CXR validation set that contain a clearly marked findings or impression section.  When creating the validation set, we filter out 13 reports that receive malformed responses from GPT-4. After also filtering out reports where the findings or impression section is missing, we arrive at a final validation set with 915 studies.

\subsection{Test Datasets}
We evaluate all models using two versions of our test set. The first is an original set of reports taken directly from MIMIC-CXR. The second is a processed set of these reports where GPT-4 has removed references to prior exams, following the procedure described above. 

We first use GPT-4 to remove references to prior exams from 2,919 reports in the MIMIC-CXR test set that contain a clearly marked findings or impression section. Unlike in the training dataset, we include reports regardless of whether they contain the substring ``compar''. When creating the processed test set, a research assistant manually fixes GPT-4's malformed responses, rather than excluding these reports. After removing reports where the findings or impression section is missing in one or both versions of the dataset, we arrive at our final test sets, each representing the same 1,383 studies.

We expect that a model that frequently refers to prior exams may better match the original set of reports, while an otherwise identical model that rarely refers to prior exams would score higher on the processed set. Since some of our models are likely to hallucinate prior exams far more frequently than others, we provide results from evaluating reports on both distributions.

\subsection{DPO Experiments}
As our first baseline, we examine our original pretrained model, without fine-tuning it further. As a second baseline, we perform supervised fine-tuning on our processed reports; this experiment essentially tests whether the DPO loss is necessary to avoid hallucinating prior exams or whether exposure to our GPT-4-processed data is sufficient. Finally, we fine-tune the model on our preference dataset using 3 DPO losses:
\begin{enumerate}
    \item $\gamma$ = 1, or the standard DPO loss. In this setting, we give equal weight to all parts of a response.
    \item $\gamma$ = .5. In this setting, we double the weight of relevant lines, compared to irrelevant ones.
    \item $\gamma$ = 0. We effectively ignore irrelevant lines, which are the same in both reports. Instead we focus only on the differences between reports.
\end{enumerate}
We fine-tune all models for 12,000 iterations. Similar to past work on DPO~\citep{rafailov2023direct}, we warm up the learning rate from 0 by 300 iterations, reaching a maximum value of 5e-7, and set $\beta = .1$. We use the RMSprop optimizer with a batch size of 16, where the weight decay is set to 0.05. In the training phase, we apply a random cropping method to process input images, with the cropped region varying between 50\% and 100\% of the original image size. Subsequently, these cropped images are resized to a standardized dimension of 224×224 pixels. We use LoRA~\citep{hu2021lora} for parameter-efficient tuning. We saved a checkpoint every 2,000 iterations. For each checkpoint, we calculated both the RadCliq-v1 score and the average number of lines with priors on the validation set. We then ranked all the checkpoints based on these two metrics. The checkpoint with the highest average ranking was selected for testing.


\begin{figure}
\includegraphics[width=\textwidth]{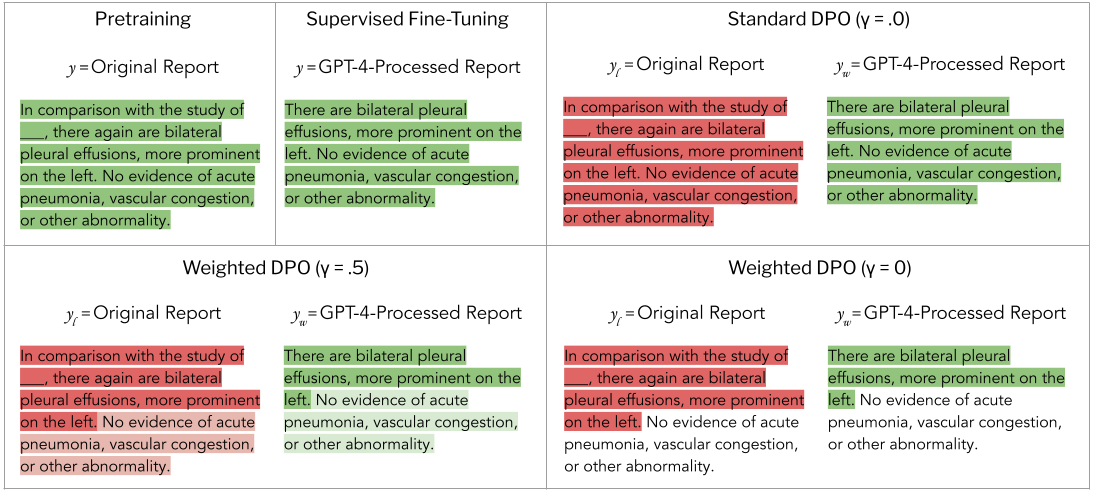}
\caption{An overview of our pretraining and fine-tuning methods. Models are rewarded for producing prior-free responses (green) and penalized for generating responses referencing prior exams (red). Depending on the DPO weighting scheme, lines unrelated to prior exams may receive less weight ($\gamma$ = .5) or be entirely ignored ($\gamma$ = 0).}
\end{figure}

\subsection{Evaluation Approach} 
When evaluating our fine-tuning approaches, we are interested in two questions: 
\begin{enumerate}
    \item Does the model produce fewer references to prior exams?
    \item Does the model maintain its original level of clinical accuracy without overfitting?
\end{enumerate}

To estimate the number of references to prior exams, we again collaborate with radiologists to produce an extensive list of key terms associated with prior exams. We measure the average number of lines in each report containing at least one of these terms to estimate the frequency of references of prior exams. We also measure the proportion of responses from each model that contains at least one of these terms. 

To determine whether the model maintains or improves on its original level of clinical accuracy, we rely on RadCliq-v1 \citep{evaluating-progress-patterns} and RadGraph-F1 \cite{jain2021radgraph}, two metrics specialized for measuring radiology report quality. We note that RadCliq-v1 is of primary importance, as it provides the best indication of radiologist opinion \citep{evaluating-progress-patterns}. We also report results on BLEU-1 and BLEU-2, two general-purpose lexical metrics. To assess significance, we calculate bootstrapped 95\% confidence intervals using scipy.stats's bootstrap function.

In addition to computing automated metrics across our entire dataset, we perform expert evaluation on a random subset of 33 studies. The 5 model responses for each study are presented in random order and annotated by either a radiologist or a radiology resident, who do not know which model produced each response. When evaluating a response, the annotator counts how many lines referred to a prior exam. They also compare the response to the original MIMIC ground-truth report, including the indication or comparison where available, and determine whether the model response contains a clinically significant discrepancy that would likely impact patient management.

\section{Results} 


\subsection{GPT-4 Annotation Quality}
We find that GPT-4 is a noisy annotator, removing approximately 60\% of references to prior exams from our training dataset. A more detailed breakdown of GPT-4's annotation performance is available in Appendix C.

\subsection{Reduction of References to Prior Exams}

\begin{table}[htbp]
\centering
\caption{Effect of our fine-tuning methods on hallucinated prior exams. We estimate the average number of lines referring to prior exams per report, as well as the proportion of reports referring to a prior exam at least once.}
\resizebox{0.75\textwidth}{!}{
\begin{tabular}{p{2.5cm}|cccccc}
\toprule
Experiment & \# Lines w/ Prior  & \% Reports w/ Prior & \\ 
\hline
Pretrained & 1.34 (1.25, 1.44) & 51.63 (48.95, 54.23) & \\ 
\hline
SFT & 1.28 (1.20, 1.37) & 55.97 (53.36, 58.64) & \\ 
\hline
$\gamma=1$  & 0.42 (0.38, 0.48) & 27.40 (25.09, 29.79) & \\ 
\hline
$\gamma=.5$  & \textbf{0.28 (0.25, 0.33)} & \textbf{20.90 (18.80, 23.07)} & \\ 
\hline
$\gamma=0$  & 0.39 (0.35, 0.44) & 25.23 (22.99, 27.62) & \\ 
\bottomrule
\end{tabular}}
\end{table}
Based on our search for key terms related to prior exams, we find that all DPO methods significantly reduce the frequency of hallucinated prior exams (Table 1 and Figure 3a). On average, our original pretrained model refers to prior exams in 1.34 lines per report, and roughly 50\% of its reports mention a prior exam at least once. In contrast, the best-performing DPO model on this metric, trained with a weighted loss where $\gamma$ = .5, refers to prior exams in .28 lines per report, significantly outperforming all of our other models. Even our worst-performing DPO model on this metric, trained with a standard DPO loss, refers to prior exams in .42 lines per report, significantly improving on both non-DPO models. Furthermore, DPO generally halves the proportion of reports that mention prior exams to 20-25\%. We find that supervised fine-tuning on processed reports does not significantly reduce hallucinated prior exams.

\begin{figure*}[t!]
    \centering
    \begin{subfigure}
        \centering
        \includegraphics[height=2in]{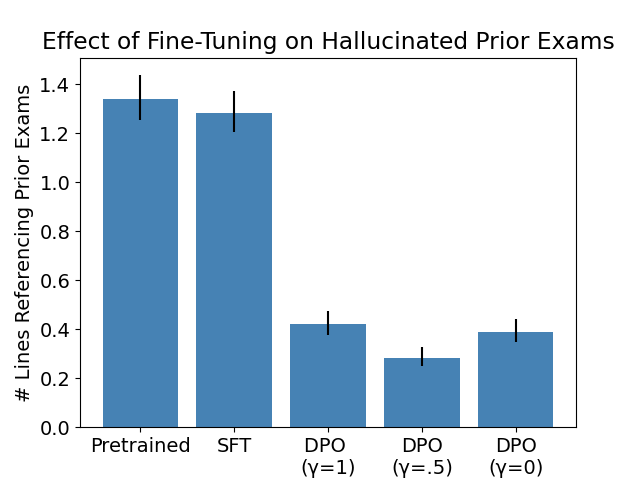}
    \end{subfigure}%
    ~ 
    \begin{subfigure}
        \centering
        \includegraphics[height=2in]{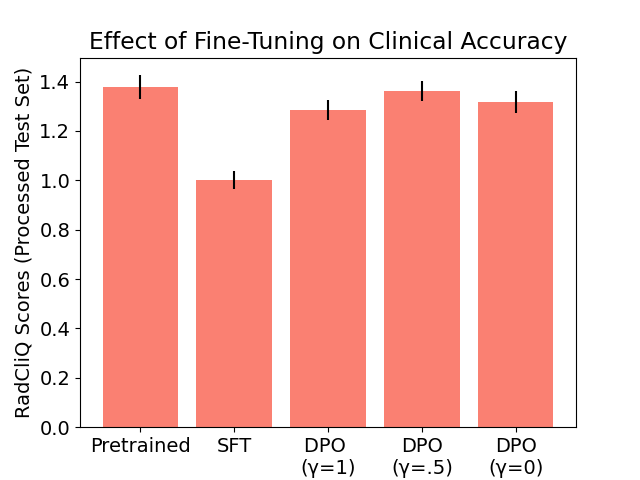}
    \end{subfigure}
    \caption{DPO fine-tuning dramatically reduces the average number of lines referring to hallucinated prior exams (left), while maintaining clinical accuracy (right). Supervised fine-tuning (SFT) improves clinical accuracy but does not meaningfully prevent the hallucination of prior exams. For both metrics, lower scores indicate higher performance.}
\end{figure*}

\subsection{Effect of Fine-Tuning on Clinical Accuracy}

\begin{table}[htbp]
\centering
\caption{Effect of our fine-tuning methods on common clinical accuracy metrics. For RadGraph-F1, higher scores indicate better performance, while for RadCliQ-V1, lower scores indicate better performance}
\resizebox{\textwidth}{!}{
\begin{tabular}{p{2cm}|cccccc}
\toprule
Experiment & \makecell{RadGraph-F1 \\(Original)} & \makecell{RadCliQ-V1 \\(Original)} & \makecell{RadGraph-F1\\ (Processed)} & \makecell{RadCliQ-V1\\ (Processed)}\\ 
\hline
Pretrained & \makecell{0.2005\\ (0.1906, 0.2115)} & \makecell{1.3914\\ (1.3413, 1.4412)} & \makecell{0.1924\\ (0.1827, 0.2032)} & \makecell{1.3782\\ (1.3295, 1.4257)} &\\ 
\hline
SFT & \makecell{0.2604\\ (0.2521, 0.2694)} & \makecell{1.0147\\ (0.9769, 1.0515)} & \makecell{0.2521\\ (0.2437, 0.2610)} & \makecell{1.0010\\ (0.9629, 1.0371)} &\\ 
\hline
$\gamma=1$ & \makecell{0.1937\\ (0.1857, 0.2016)} & \makecell{1.3352\\ (1.2951, 1.3769)} & \makecell{0.1932\\ (0.1850, 0.2012)} & \makecell{1.2847\\ (1.2455, 1.3263)} &\\ 
\hline
$\gamma=.5$  & \makecell{0.1746\\ (0.1671, 0.1824)} & \makecell{1.4187\\ (1.3780, 1.4603)} & \makecell{0.1755\\ (0.1679, 0.1834)} & \makecell{1.3623\\ (1.3225, 1.4036)} & \\ 
\hline
$\gamma=0$  & \makecell{0.1968\\ (0.1879, 0.2057)} & \makecell{1.3601\\ (1.3156, 1.4049)} & \makecell{0.1952\\ (0.1864, 0.2041)} & \makecell{1.3158\\ (1.2721, 1.3603)} &\\ 
\bottomrule
\end{tabular}}
\end{table}

We next measure the accuracy our models using specialized clinical metrics and find that the original pretrained model and DPO models perform similarly, while supervised fine-tuning consistently achieves the highest clinical accuracy (Table 2 and Figure 3b). For RadGraph-F1, higher scores indicate better performance. For RadCliq-V1, lower scores indicate better performance. For performance on lexical metrics, please see Appendix D.

\textit{RadGraph-F1.} On the original, unprocessed dataset, the original pretrained model outperforms the DPO models, though the difference is only significant when $\gamma = .5$. On the processed dataset with fewer references to prior exams, the gap in performance shrinks, and two DPO models ($\gamma = 0$ or $1$) outperform the original pretrained model, though the differences between the DPO models and the pretrained model are again non-significant.

\textit{RadCliQ-V1.} On the original, unprocessed dataset, two DPO models ($\gamma = 0$ or $1$) outperform the original pretrained model, while the third DPO model ($\gamma = .5$) performs worse; these differences are not significant. On the processed dataset, all three DPO models outperform the original pretrained model. For one of these models ($\gamma = 1$), the improvement is significant.

\subsection{Human Evaluation}

\begin{table}[htbp]
\centering
\caption{We report the average number of lines referring to prior exams per report, the proportion of reports that refer to a prior exam at least once, and the percentage reports with a clinically significant error, based on expert evaluation of 33 studies.}
\begin{tabular}{p{2cm}|cccccc}
\toprule
Experiment & \# Lines w/ Prior & \% Reports w/ Prior & \% Reports with Major Error \ \\ 
\hline
Pretrained & .97 & 33\% & 64\%\\ 
\hline
SFT & .97 & 45\% & 61\%\\ 
\hline
$\gamma=1$ & .52 & 24\% & 64\%\\ 
\hline
$\gamma=.5$ & .30 & 18\% & 85\%\\ 
\hline
$\gamma=0$ & .39 & 18\% & 70\%\\
\bottomrule
\end{tabular}
\end{table}
Though less pronounced, the trends from our expert evaluation of 33 studies were largely consistent with our findings on the whole dataset. We again find that DPO consistently reduces the number of hallucinated prior exams. Setting $\gamma= 0$ or $1$ resulted in fairly little change to clinical accuracy. Again, setting $\gamma= .5$ led to the largest reduction in hallucinated prior exams but slightly worsened clinical accuracy. Additionally, we find that our automatic metric for counting references to prior exams is generally accurate, agreeing with radiologists on over 80\% of responses with a tendency towards overcounting otherwise. More details are provided in Appendix B.


\section{Discussion} 
We find that, across all values of $\gamma$, DPO methods are effective in reducing how frequently the model refers to prior exams. Setting $\gamma = .5$ reduces lines with these unwanted references by a factor of 4.8, at the cost of slightly lower clinical accuracy. Our other two DPO models, where  $\gamma = 1$ or $0$, reduce lines with these references by a factor of 3.2 or 3.4, with higher clinical accuracy. When measured by RadCliQ-V1, a metric designed specifically to reflect radiologist opinions \citep{evaluating-progress-patterns}, these two DPO models outperform the pretrained model on both versions of our dataset. When measured by RadGraph-F1, the clinical accuracy of these two DPO models is comparable to the pretrained model's on the original dataset and slightly better on the processed dataset. 

Prior work has raised concerns that DPO's improvements may stem from improvements in fine-tuning data (i.e. from processing by strong models like GPT-4), rather than from the DPO loss itself \citep{sharma2024critical}. However, we show that data quality alone cannot reduce hallucinated prior exams. Supervised fine-tuning on our processed training dataset improves clinical accuracy, perhaps because cases mentioning prior exams tend to be especially challenging, but it does not meaningfully reduce the number of lines referencing prior exams. We therefore find that the DPO loss itself is necessary for suppressing unwanted behaviors.

\paragraph{Limitations}

We use GPT-4 to annotate our preference dataset, identifying and rewriting lines that refer to prior exams. While GPT-4 is generally capable of reducing references to prior exams, it does not eliminate them entirely, with potentially negative implications for our results. We hypothesize references remaining in our training set limit the performance of our DPO methods, so future work may find that DPO is even more effective with higher-quality annotations. Likewise, we hypothesize that performance on our test sets underrepresents the clinical accuracy of our DPO models, since even our processed test set still contains some references to prior exams. To better assess performance on CXR report generation in the future, it would be useful to develop new benchmark datasets that are entirely free of references to prior exams, better representing the ideal target distribution. 

\acks{Many thanks to Rafael Rafailov and Julian Acosta for valuable discussions on this project. We also thank the Microsoft Accelerating Foundation Models Research Grant for providing Azure compute credits.}

\bibliography{sample}

\begin{thebibliography}{30}
\providecommand{\natexlab}[1]{#1}
\providecommand{\url}[1]{\texttt{#1}}
\expandafter\ifx\csname urlstyle\endcsname\relax
  \providecommand{\doi}[1]{doi: #1}\else
  \providecommand{\doi}{doi: \begingroup \urlstyle{rm}\Url}\fi

\bibitem[Ahn et~al.(2024)Ahn, Gwon, Kim, Jun, and Park]{ahn2024note}
Imjin Ahn, Hansle Gwon, Young-Hak Kim, Tae~Joon Jun, and Sanghyun Park.
\newblock Note: Notable generation of patient text summaries through efficient approach based on direct preference optimization, 2024.

\bibitem[Amini et~al.(2024)Amini, Vieira, and Cotterell]{amini2024direct}
Afra Amini, Tim Vieira, and Ryan Cotterell.
\newblock Direct preference optimization with an offset, 2024.

\bibitem[Castricato et~al.(2024)Castricato, Lile, Anand, Schoelkopf, Verma, and Biderman]{castricato2024suppressing}
Louis Castricato, Nathan Lile, Suraj Anand, Hailey Schoelkopf, Siddharth Verma, and Stella Biderman.
\newblock Suppressing pink elephants with direct principle feedback, 2024.

\bibitem[Chaves et~al.(2024)Chaves, Huang, Xu, Xu, Usuyama, Zhang, Wang, Xie, Khademi, Yang, Awadalla, Gong, Hu, Yang, Li, Gao, Gu, Wong, Wei, Naumann, Chen, Lungren, Yeung-Levy, Langlotz, Wang, and Poon]{chaves2024training}
Juan Manuel~Zambrano Chaves, Shih-Cheng Huang, Yanbo Xu, Hanwen Xu, Naoto Usuyama, Sheng Zhang, Fei Wang, Yujia Xie, Mahmoud Khademi, Ziyi Yang, Hany Awadalla, Julia Gong, Houdong Hu, Jianwei Yang, Chunyuan Li, Jianfeng Gao, Yu~Gu, Cliff Wong, Mu~Wei, Tristan Naumann, Muhao Chen, Matthew~P. Lungren, Serena Yeung-Levy, Curtis~P. Langlotz, Sheng Wang, and Hoifung Poon.
\newblock Training small multimodal models to bridge biomedical competency gap: A case study in radiology imaging, 2024.

\bibitem[Dou et~al.(2024)Dou, Jin, Jiao, Zhao, Zhao, and Tao]{dou2024integrating}
Chengfeng Dou, Zhi Jin, Wenpin Jiao, Haiyan Zhao, Yongqiang Zhao, and Zhenwei Tao.
\newblock Integrating physician diagnostic logic into large language models: Preference learning from process feedback, 2024.

\bibitem[Endo et~al.(2021)Endo, Krishnan, Krishna, Ng, and Rajpurkar]{pmlr-v158-endo21a}
Mark Endo, Rayan Krishnan, Viswesh Krishna, Andrew~Y. Ng, and Pranav Rajpurkar.
\newblock Retrieval-based chest x-ray report generation using a pre-trained contrastive language-image model.
\newblock In Subhrajit Roy, Stephen Pfohl, Emma Rocheteau, Girmaw~Abebe Tadesse, Luis Oala, Fabian Falck, Yuyin Zhou, Liyue Shen, Ghada Zamzmi, Purity Mugambi, Ayah Zirikly, Matthew B.~A. McDermott, and Emily Alsentzer, editors, \emph{Proceedings of Machine Learning for Health}, volume 158 of \emph{Proceedings of Machine Learning Research}, pages 209--219. PMLR, 04 Dec 2021.
\newblock URL \url{https://proceedings.mlr.press/v158/endo21a.html}.

\bibitem[Gunjal et~al.(2024)Gunjal, Yin, and Bas]{gunjal2024detecting}
Anisha Gunjal, Jihan Yin, and Erhan Bas.
\newblock Detecting and preventing hallucinations in large vision language models, 2024.

\bibitem[Hu et~al.(2021)Hu, Shen, Wallis, Allen-Zhu, Li, Wang, Wang, and Chen]{hu2021lora}
Edward~J Hu, Yelong Shen, Phillip Wallis, Zeyuan Allen-Zhu, Yuanzhi Li, Shean Wang, Lu~Wang, and Weizhu Chen.
\newblock Lora: Low-rank adaptation of large language models.
\newblock \emph{arXiv preprint arXiv:2106.09685}, 2021.

\bibitem[Hyland et~al.(2024)Hyland, Bannur, Bouzid, Castro, Ranjit, Schwaighofer, Pérez-García, Salvatelli, Srivastav, Thieme, Codella, Lungren, Wetscherek, Oktay, and Alvarez-Valle]{hyland2024maira1}
Stephanie~L. Hyland, Shruthi Bannur, Kenza Bouzid, Daniel~C. Castro, Mercy Ranjit, Anton Schwaighofer, Fernando Pérez-García, Valentina Salvatelli, Shaury Srivastav, Anja Thieme, Noel Codella, Matthew~P. Lungren, Maria~Teodora Wetscherek, Ozan Oktay, and Javier Alvarez-Valle.
\newblock Maira-1: A specialised large multimodal model for radiology report generation, 2024.

\bibitem[Ivison et~al.(2023)Ivison, Wang, Pyatkin, Lambert, Peters, Dasigi, Jang, Wadden, Smith, Beltagy, and Hajishirzi]{ivison2023camels}
Hamish Ivison, Yizhong Wang, Valentina Pyatkin, Nathan Lambert, Matthew Peters, Pradeep Dasigi, Joel Jang, David Wadden, Noah~A. Smith, Iz~Beltagy, and Hannaneh Hajishirzi.
\newblock Camels in a changing climate: Enhancing lm adaptation with tulu 2, 2023.

\bibitem[Jain et~al.(2021)Jain, Agrawal, Saporta, Truong, Bui, Chambon, Zhang, Lungren, Ng, Langlotz, et~al.]{jain2021radgraph}
Saahil Jain, Ashwin Agrawal, Adriel Saporta, Steven Truong, Tan Bui, Pierre Chambon, Yuhao Zhang, Matthew~P Lungren, Andrew~Y Ng, Curtis Langlotz, et~al.
\newblock Radgraph: Extracting clinical entities and relations from radiology reports.
\newblock In \emph{Thirty-fifth Conference on Neural Information Processing Systems Datasets and Benchmarks Track (Round 1)}, 2021.

\bibitem[Jeong et~al.(2023)Jeong, Tian, Li, Hartung, Behzadi, Calle, Osayande, Pohlen, Adithan, and Rajpurkar]{jeong2023multimodal}
Jaehwan Jeong, Katherine Tian, Andrew Li, Sina Hartung, Fardad Behzadi, Juan Calle, David Osayande, Michael Pohlen, Subathra Adithan, and Pranav Rajpurkar.
\newblock Multimodal image-text matching improves retrieval-based chest x-ray report generation, 2023.

\bibitem[Johnson et~al.(2019)Johnson, Pollard, Berkowitz, Greenbaum, Lungren, Deng, Mark, and Horng]{johnson2019mimic}
Alistair~EW Johnson, Tom~J Pollard, Seth~J Berkowitz, Nathaniel~R Greenbaum, Matthew~P Lungren, Chih-ying Deng, Roger~G Mark, and Steven Horng.
\newblock Mimic-cxr, a de-identified publicly available database of chest radiographs with free-text reports.
\newblock \emph{Scientific data}, 6\penalty0 (1):\penalty0 317, 2019.

\bibitem[Kale et~al.(2023)Kale, pushpak Bhattacharyya, and Jadhav]{kale2023replace}
Kaveri Kale, pushpak Bhattacharyya, and Kshitij Jadhav.
\newblock Replace and report: Nlp assisted radiology report generation, 2023.

\bibitem[Liu et~al.(2024)Liu, Qin, Wu, Shen, Khalman, Joshi, Zhao, Saleh, Baumgartner, Liu, Liu, and Wang]{liu2024lipo}
Tianqi Liu, Zhen Qin, Junru Wu, Jiaming Shen, Misha Khalman, Rishabh Joshi, Yao Zhao, Mohammad Saleh, Simon Baumgartner, Jialu Liu, Peter~J. Liu, and Xuanhui Wang.
\newblock Lipo: Listwise preference optimization through learning-to-rank, 2024.

\bibitem[Liu et~al.(2021)Liu, Lin, Cao, Hu, Wei, Zhang, Lin, and Guo]{liu2021swin}
Ze~Liu, Yutong Lin, Yue Cao, Han Hu, Yixuan Wei, Zheng Zhang, Stephen Lin, and Baining Guo.
\newblock Swin transformer: Hierarchical vision transformer using shifted windows.
\newblock In \emph{Proceedings of the IEEE/CVF international conference on computer vision}, pages 10012--10022, 2021.

\bibitem[Miura et~al.(2021)Miura, Zhang, Tsai, Langlotz, and Jurafsky]{miura-etal-2021-improving}
Yasuhide Miura, Yuhao Zhang, Emily Tsai, Curtis Langlotz, and Dan Jurafsky.
\newblock Improving factual completeness and consistency of image-to-text radiology report generation.
\newblock In Kristina Toutanova, Anna Rumshisky, Luke Zettlemoyer, Dilek Hakkani-Tur, Iz~Beltagy, Steven Bethard, Ryan Cotterell, Tanmoy Chakraborty, and Yichao Zhou, editors, \emph{Proceedings of the 2021 Conference of the North American Chapter of the Association for Computational Linguistics: Human Language Technologies}, pages 5288--5304, Online, June 2021. Association for Computational Linguistics.
\newblock \doi{10.18653/v1/2021.naacl-main.416}.
\newblock URL \url{https://aclanthology.org/2021.naacl-main.416}.

\bibitem[Qin and Song(2022)]{qin-song-2022-reinforced}
Han Qin and Yan Song.
\newblock Reinforced cross-modal alignment for radiology report generation.
\newblock In Smaranda Muresan, Preslav Nakov, and Aline Villavicencio, editors, \emph{Findings of the Association for Computational Linguistics: ACL 2022}, pages 448--458, Dublin, Ireland, May 2022. Association for Computational Linguistics.
\newblock \doi{10.18653/v1/2022.findings-acl.38}.
\newblock URL \url{https://aclanthology.org/2022.findings-acl.38}.

\bibitem[Rafailov et~al.(2023)Rafailov, Sharma, Mitchell, Ermon, Manning, and Finn]{rafailov2023direct}
Rafael Rafailov, Archit Sharma, Eric Mitchell, Stefano Ermon, Christopher~D. Manning, and Chelsea Finn.
\newblock Direct preference optimization: Your language model is secretly a reward model, 2023.

\bibitem[Ramesh et~al.(2022)Ramesh, Chi, and Rajpurkar]{ramesh2022improving}
Vignav Ramesh, Nathan~Andrew Chi, and Pranav Rajpurkar.
\newblock Improving radiology report generation systems by removing hallucinated references to non-existent priors, 2022.

\bibitem[Ranjit et~al.(2023)Ranjit, Ganapathy, Manuel, and Ganu]{ranjit2023retrieval}
Mercy Ranjit, Gopinath Ganapathy, Ranjit Manuel, and Tanuja Ganu.
\newblock Retrieval augmented chest x-ray report generation using openai gpt models, 2023.

\bibitem[Sharma et~al.(2024)Sharma, Keh, Mitchell, Finn, Arora, and Kollar]{sharma2024critical}
Archit Sharma, Sedrick Keh, Eric Mitchell, Chelsea Finn, Kushal Arora, and Thomas Kollar.
\newblock A critical evaluation of ai feedback for aligning large language models, 2024.

\bibitem[Touvron et~al.(2023)Touvron, Martin, Stone, Albert, Almahairi, Babaei, Bashlykov, Batra, Bhargava, Bhosale, et~al.]{touvron2023llama}
Hugo Touvron, Louis Martin, Kevin Stone, Peter Albert, Amjad Almahairi, Yasmine Babaei, Nikolay Bashlykov, Soumya Batra, Prajjwal Bhargava, Shruti Bhosale, et~al.
\newblock Llama 2: Open foundation and fine-tuned chat models.
\newblock \emph{arXiv preprint arXiv:2307.09288}, 2023.

\bibitem[Tu et~al.(2023)Tu, Azizi, Driess, Schaekermann, Amin, Chang, Carroll, Lau, Tanno, Ktena, Mustafa, Chowdhery, Liu, Kornblith, Fleet, Mansfield, Prakash, Wong, Virmani, Semturs, Mahdavi, Green, Dominowska, y~Arcas, Barral, Webster, Corrado, Matias, Singhal, Florence, Karthikesalingam, and Natarajan]{tu2023generalist}
Tao Tu, Shekoofeh Azizi, Danny Driess, Mike Schaekermann, Mohamed Amin, Pi-Chuan Chang, Andrew Carroll, Chuck Lau, Ryutaro Tanno, Ira Ktena, Basil Mustafa, Aakanksha Chowdhery, Yun Liu, Simon Kornblith, David Fleet, Philip Mansfield, Sushant Prakash, Renee Wong, Sunny Virmani, Christopher Semturs, S~Sara Mahdavi, Bradley Green, Ewa Dominowska, Blaise~Aguera y~Arcas, Joelle Barral, Dale Webster, Greg~S. Corrado, Yossi Matias, Karan Singhal, Pete Florence, Alan Karthikesalingam, and Vivek Natarajan.
\newblock Towards generalist biomedical ai, 2023.

\bibitem[Xu et~al.(2023)Xu, Yang, Kelly, Sieniek, Kohlberger, Ma, Weng, Kiraly, Kazemzadeh, Melamed, Park, Strachan, Liu, Lau, Singh, Chen, Etemadi, Kalidindi, Matias, Chou, Corrado, Shetty, Tse, Prabhakara, Golden, Pilgrim, Eswaran, and Sellergren]{xu2023elixr}
Shawn Xu, Lin Yang, Christopher Kelly, Marcin Sieniek, Timo Kohlberger, Martin Ma, Wei-Hung Weng, Atilla Kiraly, Sahar Kazemzadeh, Zakkai Melamed, Jungyeon Park, Patricia Strachan, Yun Liu, Chuck Lau, Preeti Singh, Christina Chen, Mozziyar Etemadi, Sreenivasa~Raju Kalidindi, Yossi Matias, Katherine Chou, Greg~S. Corrado, Shravya Shetty, Daniel Tse, Shruthi Prabhakara, Daniel Golden, Rory Pilgrim, Krish Eswaran, and Andrew Sellergren.
\newblock Elixr: Towards a general purpose x-ray artificial intelligence system through alignment of large language models and radiology vision encoders, 2023.

\bibitem[Yan et~al.(2023)Yan, Liu, Kuo, Adithan, Reis, Kwak, Venugopal, O{'}Connell, Saenz, Rajpurkar, and Moor]{yan-etal-2023-style}
Benjamin Yan, Ruochen Liu, David Kuo, Subathra Adithan, Eduardo Reis, Stephen Kwak, Vasantha Venugopal, Chloe O{'}Connell, Agustina Saenz, Pranav Rajpurkar, and Michael Moor.
\newblock Style-aware radiology report generation with {R}ad{G}raph and few-shot prompting.
\newblock In Houda Bouamor, Juan Pino, and Kalika Bali, editors, \emph{Findings of the Association for Computational Linguistics: EMNLP 2023}, pages 14676--14688, Singapore, December 2023. Association for Computational Linguistics.
\newblock \doi{10.18653/v1/2023.findings-emnlp.977}.
\newblock URL \url{https://aclanthology.org/2023.findings-emnlp.977}.

\bibitem[Ye et~al.(2023)Ye, Liu, Chong, Zhou, Hua, and Liu]{ye2023qilinmed}
Qichen Ye, Junling Liu, Dading Chong, Peilin Zhou, Yining Hua, and Andrew Liu.
\newblock Qilin-med: Multi-stage knowledge injection advanced medical large language model, 2023.

\bibitem[Yu et~al.(2023)Yu, Endo, Krishnan, Pan, Tsai, Reis, Fonseca, Lee, Abad, Ng, Langlotz, Venugopal, and Rajpurkar]{evaluating-progress-patterns}
Feiyang Yu, Mark Endo, Rayan Krishnan, Ian Pan, Andy Tsai, Eduardo~Pontes Reis, Eduardo Kaiser Ururahy~Nunes Fonseca, Henrique Min~Ho Lee, Zahra Shakeri~Hossein Abad, Andrew~Y. Ng, Curtis~P. Langlotz, Vasantha~Kumar Venugopal, and Pranav Rajpurkar.
\newblock Evaluating progress in automatic chest x-ray radiology report generation.
\newblock 2023.
\newblock URL \url{https://doi.org/10.1016/j.patter.2023.100802}.

\bibitem[Zhou et~al.(2024{\natexlab{a}})Zhou, Cui, Rafailov, Finn, and Yao]{zhou2024aligning}
Yiyang Zhou, Chenhang Cui, Rafael Rafailov, Chelsea Finn, and Huaxiu Yao.
\newblock Aligning modalities in vision large language models via preference fine-tuning, 2024{\natexlab{a}}.

\bibitem[Zhou et~al.(2024{\natexlab{b}})Zhou, Shi, Wei, Alabi, Yue, and Vercauteren]{zhou2024large}
Zijian Zhou, Miaojing Shi, Meng Wei, Oluwatosin Alabi, Zijie Yue, and Tom Vercauteren.
\newblock Large model driven radiology report generation with clinical quality reinforcement learning, 2024{\natexlab{b}}.

\end{thebibliography}

\newpage
\appendix
\section*{Appendix A. GPT-4 Prompt for Removing References to Prior Exams.}

We dynamically created few-shot prompts for each report, depending on which common ``prior exam keywords'' appeared in it. We divide reports into sentences by splitting on this substring: ``. ''. All prompts began with the following prefix:

\begin{minipage}{\linewidth}

\begin{verbatim}
Instructions: Return only a json object for this radiology report, with 
a key-value pair for every line.
Each line starts with a numerical id. The key will be the id. The value 
will be another JSON object. 
Inside the value object, set the ``prior_cat'' attribute to say whether 
this line makes a comparison to a prior exam. 
``prior_cat'' must take one of three possible values: ``none'', 
``partial'', ``all''.
1.) If the sentence has some clear information about the current exam, 
set ``prior_cat'' as ``partial'' and then add a ``partial_rewrite'' 
attribute to the value JSON object. For the value of ``partial_rewrite'', 
rewrite the sentence to keep only that information, without any 
comparison to a prior report. 
2.) If there is no comparison, set ``prior_cat'' as ``none''. Do not 
rewrite the sentence.
3.) In the rare case that the sentence has absolutely no clear 
information about the current exam (e.g. does not mention that any 
abnormality is present or absent), set ``prior_cat'' as ``all''. Do not 
rewrite the sentence.
Here is an example:
Report: [0] No acute pulmonary process. [1] No significant changes from 
last exam.
JSON: {``0'':{``prior_cat'':``none''}, ``1'':{``prior_cat'':``all''}}
Examples of individual lines and their value objects: 
Heart is enlarged. -> {``prior_cat'':``none''}
Heart size is stable. -> {``prior_cat'':``all''}
In comparison with the study on ___, heart has enlarged, while lungs 
remain clear. -> {``prior_cat'': ``partial'', ``partial_rewrite'': 
``Heart is enlarged, while lungs are clear''.}
\end{verbatim}
\end{minipage}

Additionally, we identified 37 other terms that commonly appear in references to prior exams and, in collaboration with radiologists, provided 1-2 examples showing how to handle that keyword in different circumstances. For instance, here are the examples for the keyword ``stable'':

\begin{minipage}{\linewidth}

\begin{verbatim}
 ``stable'': ``Heart size and mediastinum are stable -> {``prior_cat'':
 ``all''}
 The cardiomediastinal silhouettes are stable reflective of a tortuous 
 thoracic aorta. -> {``prior_cat'':``partial'',  ``partial_rewrite'': 
 ``The cardiomediastinal silhouettes are reflective of a tortuous 
 thoracic aorta.''}''
\end{verbatim}
\end{minipage}

For every term that appeared in a report, we appended the corresponding examples to the prompt.

\section*{Appendix B. Estimating Number of Lines Referencing Prior Exams.}
To estimate how many lines in each report reference a prior exam, we split each report into sentences. Using case-insensitive matching, we then count how many sentences contain one of the following 43 substrings, which commonly appear in references to prior exams:
\begin{quotation}
`more', `regress', `advanc', `less', `fewer', `constant', 
`unchanged', `prior', `new', `stable', `progressed', `interval', `previous', `further', `again', `since', `increase', `improve', `remain', `worse', `persist', `remov', `similar', `cleared', `earlier', `existing', `decrease', `reduc', `recurren', `redemonstrat', `resol', `still', `has enlarged', `lower', `larger', `extubated', `smaller', `higher', `continue', 'compar', 'change',`develop', `before'
\end{quotation}

We use the results from our expert evaluation to validate our automatic metric. We find that the automatic metric and the radiologist count find the same number of lines with references to prior exams for 135 out of 165 AI-generated reports. For the remaining 30 reports, our automatic metric tends to overcount the number of references. Compared to radiologists, the automatic metric finds more references to prior exams in 22 reports; across all reports, it finds an average of .76 references to prior exams per report, compared to .63 from radiologists. The mean absolute difference between the automatic metric and radiologist counts is .23.

\section*{Appendix C. Effect of GPT-4 Annotation on References to Prior Exams.}
When using GPT-4 to remove references to prior exams, we find that GPT-4 reduces but does not fully eliminate these unwanted references. It most commonly applies the label ``none'', indicating that a line has no reference to a prior. The second most common label is ``partial''. It rarely uses ``all'', indicating that most lines have some useful information that can be preserved even when removing references to prior exams.


\begin{table}[htbp]
\centering
\caption{GPT-4 label frequencies for lines in the training, validation and test sets, followed by the total number of lines in each dataset before GPT-4 processing.}
\begin{tabular}{p{2.5cm}ccc|c}
\toprule
Dataset & None & Partial & All & Total \\
\midrule
Training & 58178 & 37591 & 11468 & 107237  \\
\hline
Validation & 5077 & 1174 & 373 & 6624 \\
\hline
Test & 7426 & 3024 & 998 & 11448 \\

\bottomrule
\end{tabular}
\end{table}

To assess GPT-4's performance on this annotation task, we estimate the number of lines that refer to prior exams in our datasets, before and after GPT-4 processing. This analysis includes the 19,806 studies in our training set, 915 studies in our validation set, and 1,383 studies in our test set.

We find that lines labeled ``partial'' or ``all'' almost always refer to a prior exam, as desired. However, a substantial portion of lines labeled ``none'' also appear to refer to a prior exam, particularly in the training set. Additionally, lines labeled ``partial'' sometimes continue to refer to prior exams, even after being rewritten. Based on manual review, we find that, when a line refers to a prior exam multiple times, GPT-4 sometimes removes some but not all of the references. For example, ``Similar to findings as compared to prior chest radiograph, perhaps increased opacity along the left lateral mid to lower lung'' is rewritten as ``Findings show increased opacity along the left lateral mid to lower lung''; the term ``increased" still refers to a prior exam. We therefore find that GPT-4 provides noisy annotations, removing over half of all references to prior exams but not eliminating them entirely.

\begin{table}[htbp]
\centering
\caption{The first three columns show the estimated number of lines from each label category that refer to prior exams, before and after GPT-4 processing. The final column shows the total number of lines that refer to prior exams, before and after GPT-4 processing.}
\begin{tabular}{p{4cm}ccc|cc}
\toprule
Dataset & None & Partial & All & Total \\
\midrule
Training (Original) & 10583 & 36627 & 11302 & 58512  \\
Training (Processed) & 10583 & 12586 & 0 & \textbf{23169}  \\
\hline
Validation (Original) & 496 & 1083 & 362 & 1941  \\
Validation (Processed) & 496 & 291 & 0 & \textbf{787}  \\
\hline
Test (Original) & 915 & 2854 & 980 & 4749 \\
Test (Processed) & 915 & 802 & 0 & \textbf{1717} \\

\bottomrule
\end{tabular}
\end{table}

\section*{Appendix D. Lexical Metrics for Clinical Accuracy.}

\begin{table}[htbp]
\centering
\caption{Comparison of Fine-Tuning Methods, based on lexical similarity. For both metrics, higher scores indicate better performance.}
\begin{tabular}{p{2cm}|cccccc}
\toprule

Experiment & \makecell{Bleu-1\\(Original)} & \makecell{Bleu-2 \\(Original)} & \makecell{Bleu-1\\(Processed)} & \makecell{Bleu-2\\(Processed)} &\\ 

\hline
pretrained & \makecell{0.2394\\ (0.2283, 0.2509)} & \makecell{0.1745\\ (0.1656, 0.1842)} & \makecell{0.2532\\ (0.2417, 0.2650)} & \makecell{0.1816\\ (0.1726, 0.1915)}\\ 
\hline
sft & \makecell{0.3058\\ (0.2967, 0.3152)} & \makecell{0.2195\\ (0.2119, 0.2277)} & \makecell{0.3305\\ (0.3215, 0.3399)} & \makecell{0.2353\\ (0.2274, 0.2434)} \\ 
\hline
k1 & \makecell{0.2033\\ (0.1944, 0.2121)} & \makecell{0.1439\\ (0.1369, 0.1510)} & \makecell{0.2388\\ (0.2293, 0.2482)} & \makecell{0.1685\\ (0.1610, 0.1761)} & \\ 
\hline

k.5 & \makecell{0.1869\\ (0.1784, 0.1955)} & \makecell{0.1302\\ (0.1237, 0.1368)} & \makecell{0.2207\\ (0.2113, 0.2299)} & \makecell{0.1535\\ (0.1463, 0.1607)}\\ 
\hline
k0 & \makecell{0.1998\\ (0.1907, 0.2094)} & \makecell{0.1437\\ (0.1365, 0.1513)} & \makecell{0.2316\\ (0.2217, 0.2417)} & \makecell{0.1656\\ (0.1578, 0.1738)}\\ 

\bottomrule
\end{tabular}
\end{table}

When measuring lexical similarity using general-purpose, non-clinical metrics, the pretrained model outperforms all DPO methods on both datasets. The difference in BLEU-1 scores is not significant on the processed dataset for two of our DPO models ($\gamma = 0$ or $1$). Otherwise, the differences are significant. 

We note that all BLEU scores increased when shifting from the original to the processed dataset. This finding can be explained by the fact that our original MIMIC test set refers to prior exams even more frequently than our pretrained model does, so the processed dataset represents a distribution of reports that more closely matches our models. We hypothesize that removing even more references to prior exams from the test set would lead to further increases in the BLEU scores of our DPO models.


\end{document}